\documentclass[letterpaper]{article} 
\usepackage{aaai25}  
\usepackage{times}  
\usepackage{helvet}  
\usepackage{courier}  
\usepackage[hyphens]{url}  
\usepackage{graphicx} 
\usepackage{natbib}  
\usepackage{caption} 
\usepackage{algorithm}
\usepackage{algorithmic}
\usepackage{colortbl}
\usepackage{pifont}
\usepackage{amsmath}
\usepackage{multirow}
\usepackage{colortbl}
\usepackage[table]{xcolor}
\definecolor{lightgray}{gray}{0.9}
\definecolor{darkgreen}{RGB}{0,128,0}
\definecolor{darkred}{RGB}{139,0,0}
\definecolor{skyblue}{RGB}{135,206,235}

\newcommand{\cmark}{\textcolor{darkgreen}{\ding{51}}} 
\newcommand{\xmark}{\textcolor{darkred}{\ding{55}}}   

\usepackage{newfloat}
\usepackage{listings}
\usepackage{amsfonts}
\usepackage{ragged2e}
\DeclareCaptionStyle{ruled}{labelfont=normalfont,labelsep=colon,strut=off} 
\floatstyle{ruled}
\newfloat{listing}{tb}{lst}{}
\floatname{listing}{Listing}
\title{AI-generated Image Quality Assessment in Visual Communication}

\author {
    Yu Tian\textsuperscript{\rm 1},
    Yixuan Li\textsuperscript{\rm 1},
    Baoliang Chen\textsuperscript{\rm 2},
    Hanwei Zhu\textsuperscript{\rm 1},
    Shiqi Wang$^{*}$\textsuperscript{\rm 1},
    Sam Kwong\thanks{Corresponding authors: Shiqi Wang, Sam Kwong.}\textsuperscript{\rm 3} 
}
\affiliations {
    \textsuperscript{\rm 1}City University of Hong Kong, Hong Kong SAR, China\\
    \textsuperscript{\rm 2}South China Normal University, Guangzhou, China\\
    \textsuperscript{\rm 3}Lingnan University, Hong Kong SAR, China\\
    
    \{ytian73-c, hanwei.zhu\}@my.cityu.edu.hk, yixuanli423@gmail.com, 
    blchen@scnu.edu.cn,
    shiqwang@cityu.edu.hk,
    samkwong@ln.edu.hk
}

\usepackage{bibentry}

\begin{document}
\maketitle
\begin{abstract}
Assessing the quality of artificial intelligence-generated images (AIGIs) plays a crucial role in their application in real-world scenarios. However, traditional image quality assessment (IQA) algorithms primarily focus on low-level visual perception, while existing IQA works on AIGIs overemphasize the generated content itself, neglecting its effectiveness in real-world applications. To bridge this gap, we propose \textbf{AIGI-VC}, a quality assessment database for \textbf{AI}-\textbf{G}enerated \textbf{I}mages in \textbf{V}isual \textbf{C}ommunication, which studies the communicability of AIGIs in the advertising field from the perspectives of information clarity and emotional interaction. The dataset consists of 2,500 images spanning 14 advertisement topics and 8 emotion types. It provides coarse-grained human preference annotations and fine-grained preference descriptions, benchmarking the abilities of IQA methods in preference prediction, interpretation, and reasoning. We conduct an empirical study of existing representative IQA methods and large multi-modal models on the AIGI-VC dataset, uncovering their strengths and weaknesses.
\begin{links}
\link{Code}{https://github.com/ytian73/AIGI-VC.}
\end{links}
\end{abstract}

%

\section{Introduction}
Image generation has undergone significant advancements with the help of artificial intelligence (AI) technology~\cite{ho2020denoising,rombach2022high,bao2024improving,chen2024iterative,zhu2024adaptive}. Recent research has demonstrated the potential benefits of AI in various visual communication fields, particularly in advertising~\cite{campbell2022preparing,quan2023big,ford2023ai,akhtar2023ai}. For example, Coca-Cola used an AI platform to create a series of advertisements (ads) for its brand, creating deeper engagement than existing ones. Some large e-commerce platforms, such as Amazon and Alibaba, utilize AI technology to generate personalized ad content, enhancing the efficiency of ad development and increasing the impact of the ads. For applications that require visual communication, high-quality images not only fully and clearly convey a certain message, but also evoke the inner emotion that the visual designer wants to reflect~\cite{holbrook1984role,hussain2017automatic,yang2023emoset}. However, due to hardware limitations and technical proficiency, the quality of AI-generated images (AIGIs) varies widely, necessitating refinement and filtering before distributing them to practical applications. 

There have been substantial efforts in establishing benchmarks to facilitate research on AIGI quality assessment.~\cite{lee2024holistic,10551508,duan2024pku,10463077}. 
However, these benchmarks emphasize the quality of generated content for general purposes, overlooking the effectiveness of AIGIs in real-world applications.
For practical applications in visual communication, the primary challenges in evaluating the quality of AIGIs arise from two aspects: 1) information clarity: each element in the text message must be present and clearly depicted in the image; 2) emotional interaction: the image must powerfully evoke the intended emotion in the viewers. It is crucial to develop an IQA benchmark that is closely aligned with practical use cases.
\begin{figure}[t]
\centering
\includegraphics[width=1\linewidth]{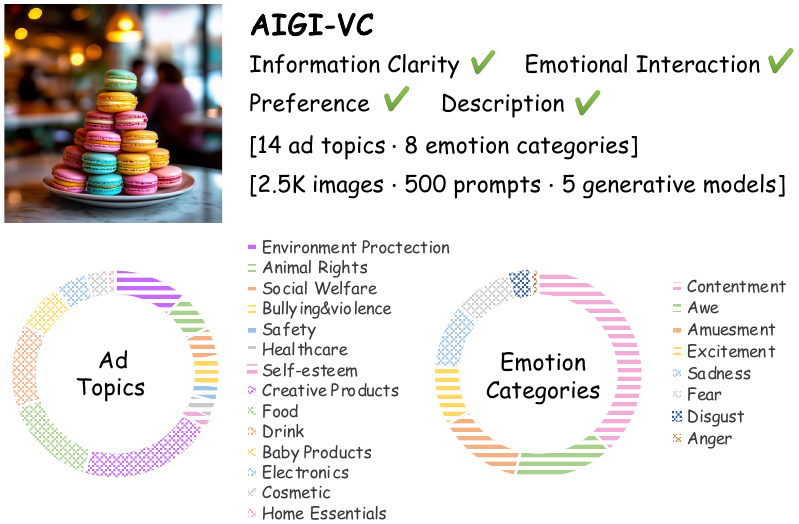}\\
\caption{Outline of the AIGI-VC dataset.}
\label{fig:dataset-overview}
\end{figure}
\begin{table*}[ht]
  \centering
\small
\setlength{\tabcolsep}{1mm}
\begin{tabular}{cccccccccc}
\hline
\multirow{3}{*}{Name} & \multicolumn{7}{c}{Evaluation Dimensions}                                                                       & \multirow{3}{*}{Score} & \multirow{3}{*}{Application} \\
\cline{2-8}
~&Text-Image &Technical & \multirow{2}{*}{Rationality} & \multirow{2}{*}{Aesthetics} & \multirow{2}{*}{Fairness} & \multirow{2}{*}{Toxicity} & Emotional &                        &                            \\
~& Alignment &Quality &~ & ~ & ~ &~ & Interaction&~ &                        \\
                      \hline
HPD v2 & \cmark                    & \xmark                 & \xmark            & \cmark          & \xmark        & \xmark        & \xmark              & Preference& General \\
Pick-a-pic& \cmark                    & \xmark                 & \xmark            & \cmark          & \xmark        & \xmark        & \xmark                & Preference& General \\
SAC & \xmark                    & \xmark                 & \xmark            & \cmark          & \xmark        & \xmark        & \xmark                   &10-Point Likert& General \\
I2P & \xmark                    & \xmark                 & \xmark            & \xmark          & \xmark        & \cmark        & \xmark                  &Percentage& General \\
GenData& \xmark                    & \xmark                 & \xmark            & \xmark          & \cmark        & \xmark        & \xmark                   &Probability& General \\
SeeTRUE& \cmark                    & \xmark                 & \xmark            & \xmark          & \xmark        & \xmark        & \xmark                 &0/1& General \\
AGIN& \xmark                    & \cmark                 & \cmark            & \xmark          & \xmark        & \xmark        & \xmark                  &MOS& General \\
AGIQA-3k& \cmark                    & \cmark                 & \cmark            & \cmark          & \xmark        & \xmark        & \xmark                &MOS& General \\
\multirow{2}{*}{ImageReward}& \multirow{2}{*}{\cmark}                    & \multirow{2}{*}{\cmark}                 & \multirow{2}{*}{\cmark}            & \multirow{2}{*}{\cmark}          & \multirow{2}{*}{\xmark}        & \multirow{2}{*}{\cmark}       &  \multirow{2}{*}{\xmark}              &5-Point Likert&\multirow{2}{*}{General} \\
~&~&~&~&~&~&~ &~ &\&Ranking& ~ \\
\multirow{2}{*}{AesMMIT}& \multirow{2}{*}{\xmark}                    & \multirow{2}{*}{\xmark}                 & \multirow{2}{*}{\xmark}            & \multirow{2}{*}{\cmark}          & \multirow{2}{*}{\xmark}        & \multirow{2}{*}{\xmark}        & \multirow{2}{*}{\cmark}       &5-Point Likert& \multirow{2}{*}{General} \\
~&~&~&~&~&~&~ &~&\&Description& ~ \\
\hline
\multirow{2}{*}{\textbf{AIGI-VC}}& \multirow{2}{*}{\cmark}                    & \multirow{2}{*}{\cmark}                 & \multirow{2}{*}{\cmark}          & \multirow{2}{*}{\cmark}          & \multirow{2}{*}{\xmark}        & \multirow{2}{*}{\xmark}        & \multirow{2}{*}{\cmark}    &Preference& Visual \\
~&~ &~ &~&~ &~&~&~&\& Description&Commnunication\\
\hline
\end{tabular}
\caption{Summary of representative AIGI databases.}
\label{tab:benchmarks}
\end{table*}
In this work, we contribute a dataset called \textbf{AIGI-VC}, the first-of-its-kind database to study the communicability of \textbf{AI}-\textbf{G}enerated \textbf{I}mages in \textbf{V}isual \textbf{C}ommunication. The overview of the AIGI-VC dataset is shown in Fig.~\ref{fig:dataset-overview}. The AIGI-VC dataset comprises a diverse collection of 2,500 images, encompassing 14 distinct ad topics and representing 8 different types of emotions.
We conduct subjective experiments via pairwise comparisons on two evaluation dimensions (i.e., information clarity and emotional interaction), collecting coarse-grained and fine-grained human preference annotations. The coarse-grained annotations provide a general sense of human preference by capturing choices between pairs of images. For the fine-grained descriptions, we provide several visual cues for each evaluation dimension as guidelines and utilize a collaborative approach between human subjects and GPT-4o~\cite{openai2023gpt4v} to collect detailed insights behind these preferences. By incorporating these annotations, AIGI-VC benchmarks the capabilities of various IQA methods in terms of preference prediction, interpretation, and reasoning.
We conduct experiments on several IQA metrics and large multi-modal models (LMMs) using the AIGI-VC dataset. Additionally, we sample three subsets from the AIGI-VC dataset to evaluate the performance of IQA metrics in handling different scenarios: AIGIs involving human-object interactions, AIGIs with fantasy content, and AIGIs evoking positive/negative emotions. Overall, we observe that the state-of-the-art models do not perform effectively when evaluating the quality of AIGIs in visual communication.

In summary, our contributions are mainly in three aspects: 1) We introduce the first-of-its-kind AIGI-VC dataset, which tackles the critical challenges of assessing the effectiveness of AIGIs in practical applications. 2) We provide human preference annotations ranging from coarse-grained to fine-grained, benchmarking the various capabilities of IQA metrics, including preference prediction, interpretation, and reasoning. 3) We perform a series of performance evaluations on state-of-the-art IQA metrics and LMMs using the AIGI-VC dataset, uncovering their relatively limited effectiveness in evaluating the communicability of AIGIs. We hope that our efforts will contribute to further advancements in the use of AIGIs for visual communication applications.
\begin{figure*}[t]
\centering
\includegraphics[width=1\linewidth]{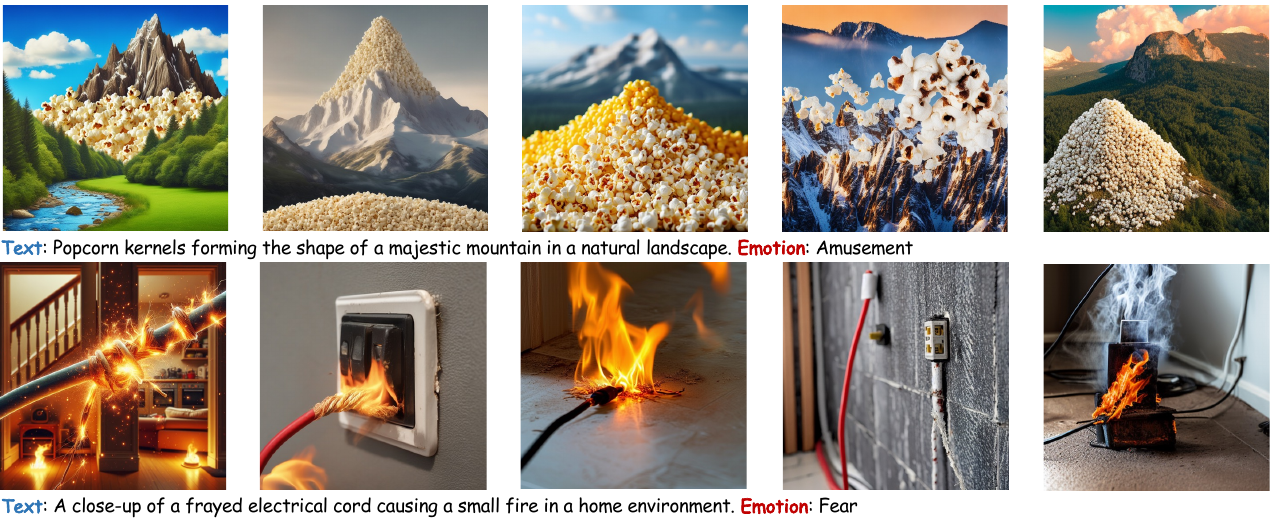}\\
\caption{Sample images from the AIGI-VC database, where the first to fifth columns show images generated by Dall$\cdot$E 3, Stable Diffusion XL, Stable Diffusion 3.0, Stable Diffusion 2.0, and Dreamlike Photoreal 2.0.}
\label{fig:samples}
\end{figure*}
\begin{figure}[t]
\centering
\includegraphics[width=0.7\linewidth]{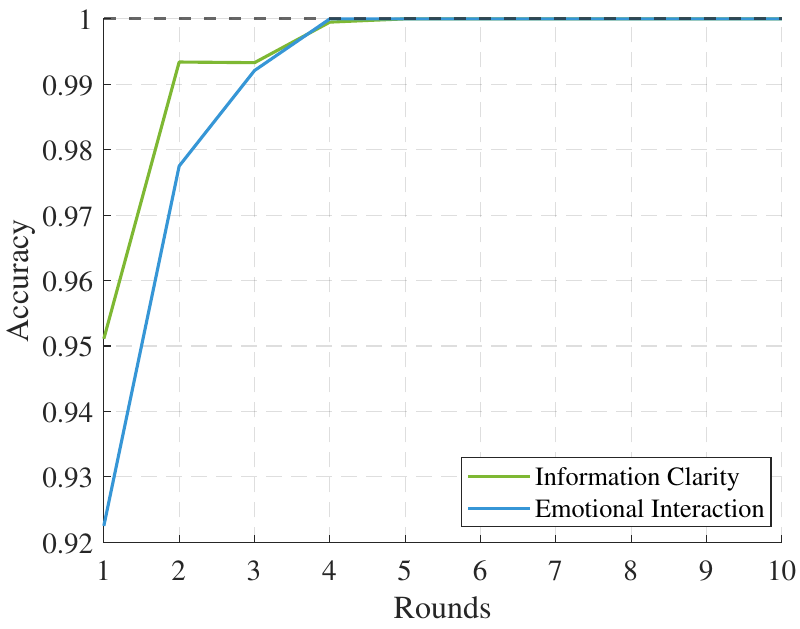}\\
\caption{Accuracy of preference choices via MAP estimation in $M$ rounds.}
\label{fig:MAP}
\end{figure}
\section{Related Works}
\subsection{Subjective Databases for AIGIs}
We present a summary of representative datasets for AIGI quality assessment in Table~\ref{tab:benchmarks}. 
Human Preference Dataset (HPDv2)~\cite{wu2023humanv2} and Pick-a-pic~\cite{kirstain2024pick} are IQA datasets for AIGIs, which focus on the overall quality in terms of text-image alignment and aesthetics. They provide binary preference choices within image pairs. 
Simulacra Aesthetic Captions (SAC)~\cite{pressmancrowson2022} dataset is designed to evaluate the aesthetics of AIGIs. It includes over 238,000 images created by GLIDE~\cite{nichol2021glide} and Stable Diffusion, with users rating their aesthetic value on a scale from 1 to 10.
Inappropriate Image Prompts (I2P)~\cite{schramowski2023safe} dataset is designed to evaluate the risk of inappropriate content in text-to-image generation tasks. It contains 4.7k prompts to produce inappropriate content.
The toxicity score is indicated by the proportion of 10 images with the same prompt classified as inappropriate by objective metrics. GenData~\cite{teo2024measuring} is designed to evaluate the fairness of generative models, which offers the probability of the sensitive attribute for each generative model. SeeTrue~\cite{yarom2024you} comprises 31,855 text-image pairs with binary annotations for alignment/misalignment. AI-Generated Image Naturalness (AGIN)~\cite{chen2023exploring} focuses on the naturalness of AIGIs from technical and rationality dimensions and provides mean opinion score (MOS) values of 6,049 images in each evaluation dimension. AGIQA-3k~\cite{li2023agiqa} contains 2,982 AIGIs with human-labeled MOS values from both perception and text-image alignment dimensions. ImageReward~\cite{xu2024imagereward} provides 137k pairs of expert comparisons, including rating and ranking from text-image alignment, fidelity, and harmlessness perspectives. Aesthetic Multi-Modality Instruction Tuning (AesMMIT)~\cite{huang2024aesexpert} studies on the aesthetic quality of AIGIs covering multiple aesthetic perception dimensions. It provides direct human feedback on aesthetic perception and understanding via progressive questions. It is worth noting that AesMMIT explores what emotion an image conveys by posing an open-ended question, rather than emphasizing whether the image effectively communicates the intended emotion.
Our proposed AIGI-VC specifically evaluates the effectiveness of AIGIs in visual communication, emphasizing information clarity and emotional interaction in practical applications.
\begin{figure}[t]
\centering
\includegraphics[width=0.7\linewidth]{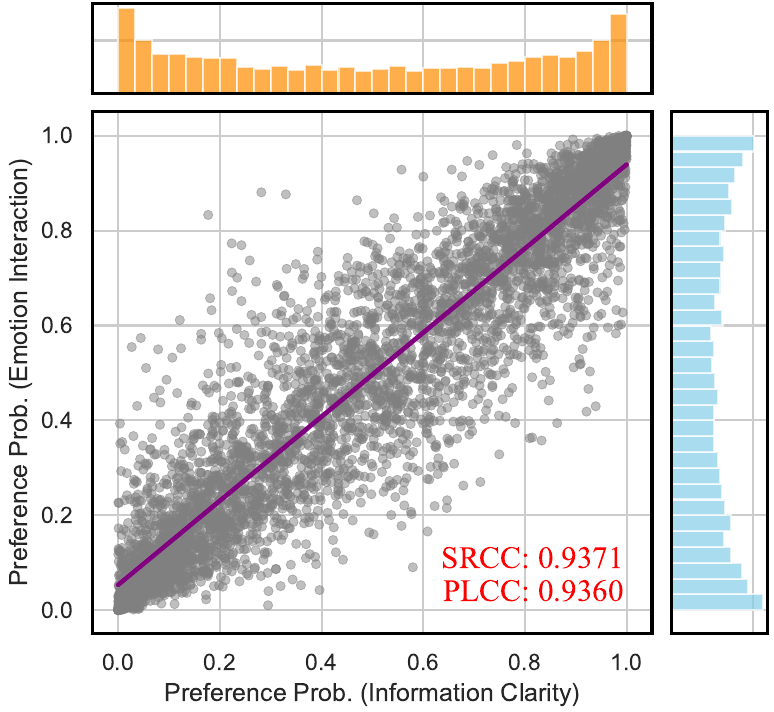}\\
\caption{Distribution of preference probabilities for image pairs in the AIGI-VC dataset.}
\label{fig:corr}
\end{figure}
\begin{figure*}[t]
\centering
\includegraphics[width=1\linewidth]{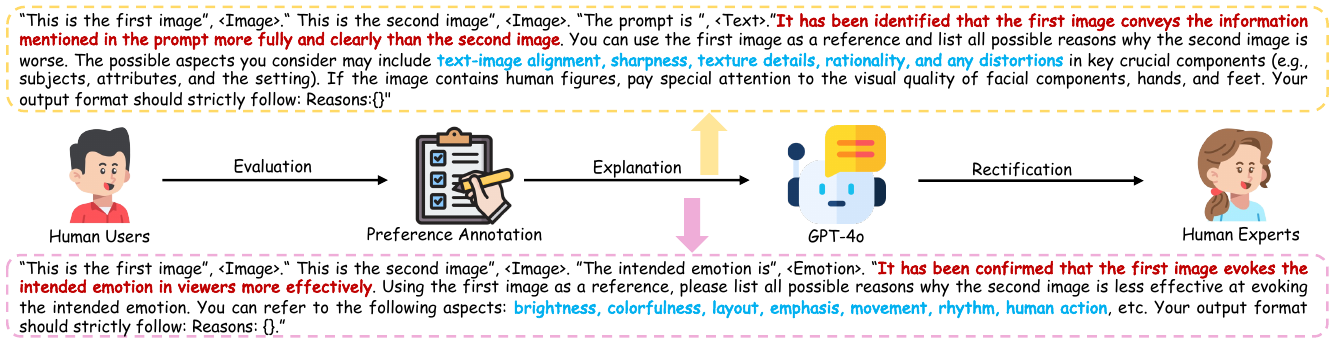}\\
\caption{The process of description generation. Given two images with preference choices collected from human users, GPT produces the initial descriptions according to visual cues influencing human preference judgments. Human experts then verify and supplement GPT-generated descriptions to produce golden descriptions.}
\label{fig:descript-process}
\end{figure*}
\subsection{Quality Assessment Metrics for AIGIs}
Existing AIGI quality assessment metrics can be roughly classified into vanilla quality assessment metrics~\cite{Gu20,10184113,10517303}, contrastive language-image pre-training (CLIP) based quality assessment metrics~\cite{hessel2021clipscore,xu2024imagereward,kirstain2024pick,li2023agiqa}, and visual question answering based quality assessment metrics~\cite{huang2024t2i,lu2024llmscore,Cho2024DSG,yarom2024you,cho2023dall,wu2024q,chen2024visual}. Typically, vanilla quality assessment metrics rely on a predefined feature extractor to derive task-specific features from the images, and the quality score is computed based on these features. However, these metrics support only single-modal input, limiting their effectiveness in evaluating the quality of AIGIs involving multimodal content. CLIP-based metrics are widely applied in text-image alignment evaluation, which measure the similarity between text and image embeddings derived from a pre-trained model capable of understanding both modalities. Recently, LMMs have exhibited exceptional linguistic capabilities across general human knowledge domains, attracting significant attention in the IQA field. In this work, we perform in-depth analyses to gain insights into the strengths and limitations of these models when evaluating the communicability of AIGIs.
\section{Dataset Construction}
\subsection{Data Collection}
To cover diverse content and emotions, the AIGI-VC dataset involves two common ad types, i.e., product ads and public service announcements (PSAs). Product ads promote commercial products or services and generally aim to evoke positive emotions to stimulate consumer interest and purchases. PSAs raise awareness about social issues and public health, often evoking negative emotions such as concern or urgency to encourage action or behavior change. According to related research on the advertising field~\cite{hussain2017automatic,sagar2024madverse}, we select 14 ad topics, where seven topics (i.e., creative products, food, drink, baby products, electronics, cosmetics, and home essentials) for product ads and seven topics (i.e., environment protection, animal rights, social welfare, safety, healthcare, and self-esteem) for PSAs. 
To streamline the design process of ads, we utilized GPT-4V as a content designer to generate diverse prompts, including textual content and intended emotions for ads based on given topics.
According to Mikels model~\cite{mikels2005emotional}, the intended emotions are selected from eight types, i.e., amusement, awe, contentment, excitement, anger, disgust, fear, and sadness. We verify and remove highly similar responses through a combination of manual review and an objective algorithm, ensuring the uniqueness and quality of the database. After this procedure, we obtain 500 distinct prompts. We employ five popular text-to-image generation models, namely Stable Diffusion XL, Stable Diffusion 2.0, Stable Diffusion 3.0~\cite{rombach2022high}, Dreamlike Photoreal 2.0~\cite{rombach2022high}, Dall$\cdot$E 3~\cite{ramesh2022hierarchical}. Ultimately, we obtain a total of 2,500 AIGIs. Each image is resized to 512$\times$512 to standardize the dataset, ensuring reducing variability related to image resolution. Some images sampled from the AIGI-VC dataset are shown in Fig.~\ref{fig:samples}.

\subsection{Human Preference Annotation}
\subsubsection{Coarse-grained Preference Choices}
We collect human opinions via the pairwise image comparison method, directly asking participants to choose their preferred image from a pair.
Generally speaking, global ranking results of $N$ test stimuli are derived from exhaustive pairwise comparisons, which involve conducting $\binom{N}{2}$ pairwise comparisons. However, this process is time-consuming and expensive. Therefore, as suggested in~\cite{zhu20242afc,prashnani2018pieapp}, we employ Thurstone’s Case V model~\cite{tsukida2011analyze} to estimate the missing human labels using a subset of the exhaustive pairwise comparison data.
Let $x_{i}$ and $x_{j}$ represent two images generated from the same prompt. We collect the preference entry $C_{i,j}$, which indicates the number of times $x_{i}$ is preferred over $x_{j}$. The global ranking scores $\mathcal{Q}=\{q_{i}\}_{i=1}^{N}$ can be estimated by solving the following 
maximum a posterior (MAP) estimation problem:
\begin{align}
    \text{arg max}_{\mathcal{Q}}&\sum_{i,j}C_{i,j}\log{(\Phi(q_{i}-q_{j}) )-\sum_{i}\frac{(q_{i})^{2}}{2}  } \notag ,\\
    \text{subject to}&\sum_{i}q_{i}=0,
\end{align}
where $\Phi(\cdot)$ is the standard normal cumulative distribution function.

To verify the reliability and effectiveness of MAP estimation, we compare the correlation between the scores estimated from a subset of the exhaustive pairwise comparison data and the true scores obtained through exhaustive pairwise comparisons. Specifically, we selected 250 images generated from 50 prompts in the AIGI-VC dataset. In each round, each image was randomly paired with another image with the same prompt. We repeated this process for 
$M$ rounds and calculated the accuracy of preference choices derived from the estimated ranking scores. Following the data reliability recommendations in~\cite{ITU,prashnani2018pieapp}, we collected responses from 20 participants (11 males and 9 females) aged 21 to 31. Due to the preference ambiguity caused by the similar quality of the two images in a pair~\cite{zhang2021uncertainty,7934456}, we focus on cases with strong estimated preferences, namely those where the preference probabilities fall outside the range of $\left[0.3, 0.7\right]$. The results are shown in Fig.~\ref{fig:MAP}, we can find that the estimated preferences can recover the true preferences when $M$ is 4. Therefore, in our subjective experiment, 20 participants are employed to label 2,000 pairs randomly sampled from the whole AIGI-VC dataset, reducing the required number of exhaustive pairwise comparisons by 60\% while producing the same preferences. We provide a visualization of the estimated preference probabilities in the AIGI-VC database, shown in Fig.~\ref{fig:corr}, from which one can observe the Spearman Rank-Order Correlation Coefficient (SRCC) and Pearson Linear Correlation Coefficient (PLCC) between the preference probabilities for information clarity and emotional interaction reach 0.9371 and 0.9360, respectively. These results demonstrate an intrinsic correlation between information clarity and emotional interaction.
\subsubsection{Fine-grained Descriptions}
We further provide detailed descriptions to determine the reasons that influence human judgments of images, enhancing the interpretability and transparency of the AIGI-VC dataset. As illustrated in Fig.~\ref{fig:descript-process}, we adopt a humans-in-the-loop strategy~\cite{wu2022survey} to reduce workload and enhance data reliability. Specifically, to obtain more detailed and comprehensive descriptions, we treat the top-ranked image in each evaluation dimension for each prompt as a pseudo-reference and incorporate GPT-4o~\cite{achiam2023gpt} to identify why the other image under the same prompt is worse than the pseudo-reference. Furthermore, we provide various visual cues for each evaluation dimension. For information clarity, the visual cues include text-image alignment, sharpness, texture details, and rationality~\cite{chen2023exploring,li2023agiqa}. For emotional interaction, the visual cues include layout, emphasis, movement, rhythm, human action, brightness, and colorfulness~\cite{10.1145/2647868.2654930,yang2023emoset}. To avoid subjective divergence, we remove image pairs where both images have similar quality. We collect responses from GPT-4o as the initial descriptions and recruit human experts to verify and supplement each GPT-generated description, creating a golden standard description. To better illustrate how those visual factors contribute to quality assessment, we present the frequently occurring words in golden standard descriptions. The results and  analyses are provided in the supplementary
materials.

\section{Evaluation on AIGI-VC}
\subsection{Experimental Settings}
\subsubsection{Baselines} We employ 14 objective metrics for performance comparisons, including one emotion classifier (WSCNet~\cite{8825564}), one vanilla quality assessment metrics designed for natural images (HyperIQA~\cite{su2020blindly}), five CLIP-based metrics tailored for AIGIs (CLIPscore~\cite{hessel2021clipscore}, AestheticScore, HPS v2~\cite{wu2023humanv2}, ImageReward~\cite{xu2024imagereward}, and PickScore~\cite{kirstain2024pick}), and seven LMMs that accept multiple images as input (mPLUG-Owl2~\cite{ye2023mplug}, LLaVA-v1.5-13B~\cite{liu2024visual}, InterLM-XC.2-vl~\cite{internlmxcomposer2}, BakLLava~\cite{BakLLaVA2024}, Idefics2~\cite{laurenccon2024matters}, Qwen-VL~\cite{Qwen-VL}, and GPT-4o. Detailed information of these LMMs is summarized in supplementary materials. It is worth noting that we re-train the WSCNet from scratch on a large-scale visual emotion dataset~\cite{yang2023emoset}. To ensure fairness, we use the default hyperparameters provided by the original models. 

\begin{table}[ht]
\centering
\small
\setlength{\tabcolsep}{1.5mm}
\begin{tabular}{c|c|cc|cc}
\hline
\multirow{2}{*}{Model}&\multirow{2}{*}{Criteria}                                                     & \multicolumn{2}{c|}{IC} & \multicolumn{2}{c}{EI}\\
\cline{3-6}
   &~&$D_{all}$& $D_{sub}$&$D_{all}$&$D_{sub}$\\
   \hline
\multirow{2}{*}{HyperIQA} &\cellcolor{lightgray}$\alpha$$\uparrow$& \cellcolor{lightgray}0.5438&\cellcolor{lightgray}0.5497&\cellcolor{lightgray}0.5404&\cellcolor{lightgray}0.5571\\
~&$\rho$$\uparrow$&0.0935&0.1085&0.1012&0.1244\\
\hline
\multirow{2}{*}
{WSCNet}&\cellcolor{lightgray}$\alpha \uparrow$&\cellcolor{lightgray}-&\cellcolor{lightgray}-&\cellcolor{lightgray}0.5366&\cellcolor{lightgray}0.5521\\
&$\rho \uparrow$&-&-&-&-\\
\hline
\multirow{2}{*}
{CLIPScore}&$\cellcolor{lightgray}\alpha$$\uparrow$&\cellcolor{lightgray}0.5654&\cellcolor{lightgray}0.5880&\cellcolor{lightgray}0.5988&\cellcolor{lightgray}0.6273\\
&$\rho \uparrow$&0.1854&0.2212&0.2659&0.3153\\
\hline
\multirow{2}{*}
{AestheticScore}&\cellcolor{lightgray}$\alpha$$\uparrow$&\cellcolor{lightgray}0.6794&\cellcolor{lightgray}0.7267&\cellcolor{lightgray}0.6832&\cellcolor{lightgray}0.7425\\
&$\rho \uparrow$&0.4731&0.5364&0.4832&0.5625\\
\hline
\multirow{2}{*}
{HPSv2}&\cellcolor{lightgray}$\alpha$$\uparrow$&\cellcolor{lightgray}0.7386&\cellcolor{lightgray}0.8131&\cellcolor{lightgray}\textbf{0.7036}&\cellcolor{lightgray}\textbf{0.7649}\\
&$\rho$$\uparrow$&0.6101&0.6612&\textbf{0.5349}&\textbf{0.5937}\\
\hline
\multirow{2}{*}
{ImageReward}&\cellcolor{lightgray}$\alpha$$\uparrow$&\cellcolor{lightgray}\underline{0.7484}&\cellcolor{lightgray}\underline{0.8227}&\cellcolor{lightgray}\underline{0.6924}&\cellcolor{lightgray}0.7481 \\
&$\rho \uparrow$&\underline{0.6709}&\underline{0.7220}&0.4687&0.5365\\
\hline
\multirow{2}{*}
{PickScore}&\cellcolor{lightgray}$\alpha \uparrow$&\cellcolor{lightgray}\textbf{0.7518}&\cellcolor{lightgray}\textbf{0.8306}&\cellcolor{lightgray}0.6912&\cellcolor{lightgray}\underline{0.7513}\\
&$\rho \uparrow$&\textbf{0.6807}&\textbf{0.7554}&\underline{0.5157}&\underline{0.5883}\\
\hline     
\end{tabular}
\caption{Comparison of IQA metrics in \textbf{preference prediction}. IC: Information clarity. EI: Emotional interaction. The best two results are highlighted in bold and underlined.}
\label{tab:preferencepred1}
\end{table}
\begin{table}[h]
\centering
\small
\setlength{\tabcolsep}{1.3mm}
\begin{tabular}{c|c|cc|cc}
\hline
\multirow{2}{*}{Model}&\multirow{2}{*}{Criteria}                                                     & \multicolumn{2}{c|}{IC} & \multicolumn{2}{c}{EI}\\
\cline{3-6}
   &   &$D_{all}$& $D_{sub}$&$D_{all}$&$D_{sub}$\\
   \hline
\multirow{2}{*}
{LLaVA-v1.5-13B}&\cellcolor{lightgray}$\alpha \uparrow$&\cellcolor{lightgray}0.4846&\cellcolor{lightgray}0.4878&\cellcolor{lightgray}0.4984&\cellcolor{lightgray}0.5083\\
&$\kappa \uparrow$&0.0296&0.0310&0.2660&0.2679\\
\hline
\multirow{2}{*}
{BakLLava}&\cellcolor{lightgray}$\alpha \uparrow$&\cellcolor{lightgray}0.4916&\cellcolor{lightgray}0.4911&\cellcolor{lightgray}0.4946&\cellcolor{lightgray}0.4882\\
&$\kappa \uparrow$&0.1634&0.1669&0.2014&0.2025\\
\hline
\multirow{2}{*}
{mPLUG-Owl2}&\cellcolor{lightgray}$\alpha \uparrow$&\cellcolor{lightgray}0.4800&\cellcolor{lightgray}0.4714&\cellcolor{lightgray}0.4846&\cellcolor{lightgray}0.4834\\
&$\kappa \uparrow$&\underline{0.4846}&\underline{0.4883}&\underline{0.4622}&\underline{0.4577}\\
\hline
\multirow{2}{*}
{IDEFICS-Instruct}&\cellcolor{lightgray}$\alpha \uparrow$&\cellcolor{lightgray}\underline{0.5524}&\cellcolor{lightgray}\underline{0.5736}&\cellcolor{lightgray}\underline{0.5902}&\cellcolor{lightgray}\underline{0.6178}\\
&$\kappa \uparrow$&0.2048&0.2102&0.3484&0.3609\\
\hline
\multirow{2}{*}
{Qwen-VL-Chat}&\cellcolor{lightgray}$\alpha \uparrow$&\cellcolor{lightgray}0.4940&\cellcolor{lightgray}0.4945&\cellcolor{lightgray}0.2760&\cellcolor{lightgray}0.2768\\
&$\kappa \uparrow$&0.0030&0.0031&0.0134&0.0148\\
\hline
\multirow{2}{*}
{InternLM-XC.2-vl}&\cellcolor{lightgray}$\alpha \uparrow$&\cellcolor{lightgray}0.3240&\cellcolor{lightgray}0.3282&\cellcolor{lightgray}0.4568&\cellcolor{lightgray}0.4565\\
&$\kappa \uparrow$&0.2010&0.2055&0.2636&0.2741\\
\hline
\multirow{2}{*}
{GPT-4o}&\cellcolor{lightgray}$\alpha \uparrow$&\cellcolor{lightgray}\textbf{0.7928}&\cellcolor{lightgray}\textbf{0.8826}&\cellcolor{lightgray}\textbf{0.7236}&\cellcolor{lightgray}\textbf{0.7993}\\
&$\kappa \uparrow$&\textbf{0.8687}&\textbf{0.9013}&\textbf{0.6424}&\textbf{0.6552}\\
\hline     
\end{tabular}
\caption{Comparison of LMMs in \textbf{preference prediction}. IC: Information clarity. EI: Emotional interaction. The best two results are highlighted in bold and underlined, respectively.}
\label{tab:preferencepred2}
\end{table}
\begin{table*}[ht]
\centering
\small
\setlength{\tabcolsep}{1.5mm}
\begin{tabular}{c|c|cccc|cccc|cccc}
\hline
\multirow{2}{*}{Model}                                                     & \multirow{2}{*}{Criteria} & \multicolumn{4}{c|}{Information Clarity} &\multicolumn{4}{c|}{Emotional Interaction}&\multicolumn{4}{c}{Overall}\\
\cline{3-14}
   &                       &I&II&III-(P)&III-(N)&I& II&III-(P)&III-(N)&I&II&III-(P)&III-(N)\\
   \hline
\multirow{2}{*}{HyperIQA} &\cellcolor{lightgray}$\alpha \uparrow$&\cellcolor{lightgray}0.5489&\cellcolor{lightgray}0.5789&\cellcolor{lightgray}0.5455&\cellcolor{lightgray}0.5430&\cellcolor{lightgray}0.5333&\cellcolor{lightgray}0.5526&\cellcolor{lightgray}0.5448& \cellcolor{lightgray}0.5401&\cellcolor{lightgray}0.5411&\cellcolor{lightgray}0.5658&\cellcolor{lightgray}0.5452&\cellcolor{lightgray}0.5416\\
&$\rho \uparrow$&0.1031&0.1546&0.0926&0.0957&0.1096&0.1590&0.1078&0.1014&0.1064&0.1568&0.1002&0.0986\\
   \hline
\multirow{2}{*}{CLIPScore} &\cellcolor{lightgray}$\alpha \uparrow$&\cellcolor{lightgray}0.5400&\cellcolor{lightgray}0.6316&\cellcolor{lightgray}0.5652&\cellcolor{lightgray}0.5646&\cellcolor{lightgray}0.5793&\cellcolor{lightgray}0.6842&\cellcolor{lightgray}0.6020&\cellcolor{lightgray}0.6008&\cellcolor{lightgray}0.5597&\cellcolor{lightgray}0.6579&\cellcolor{lightgray}0.5836&\cellcolor{lightgray}0.5827 \\
&$\rho \uparrow$&0.1291&0.3700&0.1805&0.1762&0.2123&0.4693&0.2687&0.2753&0.1707&0.4197&0.2246&0.2258\\
   \hline
\multirow{2}{*}{AestheticScore} &\cellcolor{lightgray}$\alpha \uparrow$&\cellcolor{lightgray}0.6830&\cellcolor{lightgray}0.6711&\cellcolor{lightgray}0.6904&\cellcolor{lightgray}0.6854&\cellcolor{lightgray}0.6889&\cellcolor{lightgray}0.7237&\cellcolor{lightgray}0.6988&\cellcolor{lightgray}0.6873&\cellcolor{lightgray}0.6860&\cellcolor{lightgray}0.6974&\cellcolor{lightgray}0.6946&\cellcolor{lightgray}0.6864 \\
&$\rho \uparrow$&0.4697&0.5394&0.4861&0.4844&0.4672&\underline{0.6212}&0.5028&0.4919&0.4685&0.5803&0.4945&0.4882\\
\hline
\multirow{2}{*}{HPSv2} &\cellcolor{lightgray}$\alpha \uparrow$&\cellcolor{lightgray}\underline{0.7363}&\cellcolor{lightgray}0.7368&\cellcolor{lightgray}\underline{0.7524}&\cellcolor{lightgray}0.7390&\cellcolor{lightgray}\textbf{0.7230}&\cellcolor{lightgray}\underline{0.7500}&\cellcolor{lightgray}\textbf{0.7216}&\cellcolor{lightgray}\textbf{0.7050}&\cellcolor{lightgray}\textbf{0.7297}&\cellcolor{lightgray}0.7434&\cellcolor{lightgray}\textbf{0.7370}&\cellcolor{lightgray}0.7220 \\
&$\rho \uparrow$&0.6276&0.6338&0.6141&0.6028&\textbf{0.5542}&\textbf{0.6351}&\textbf{0.5504}&\textbf{0.5445}&\textbf{0.5909}&\underline{0.6345}&0.5823&0.5737\\
\hline
\multirow{2}{*}{ImageReward} &\cellcolor{lightgray}$\alpha \uparrow$&\cellcolor{lightgray}0.7319&\cellcolor{lightgray}\textbf{0.8421}&\cellcolor{lightgray}\textbf{0.7548}&\cellcolor{lightgray}\underline{0.7500}&\cellcolor{lightgray}0.6852&\cellcolor{lightgray}0.7237&\cellcolor{lightgray}\underline{0.7000}&\cellcolor{lightgray}0.6965&\cellcolor{lightgray}0.7086&\cellcolor{lightgray}\textbf{0.7829}&\cellcolor{lightgray}\underline{0.7274}&\cellcolor{lightgray}\underline{0.7233} \\
&$\rho \uparrow$&\underline{0.6367}&\textbf{0.7335}&\textbf{0.6761}&\underline{0.6765}&0.4187&0.5950&0.4988&0.4797&0.5277&\textbf{0.6643}&\underline{0.5875}&\underline{0.5781}\\
\hline
\multirow{2}{*}{PickScore} &\cellcolor{lightgray}$\alpha \uparrow$&\cellcolor{lightgray}\textbf{0.7385}&\cellcolor{lightgray}\underline{0.7500}&\cellcolor{lightgray}0.7514&\cellcolor{lightgray}\textbf{0.7576}&\cellcolor{lightgray}\underline{0.6941}&\cellcolor{lightgray}\textbf{0.7895}&\cellcolor{lightgray}0.6974&\cellcolor{lightgray}\underline{0.6980}&\cellcolor{lightgray} \underline{0.7163}&\cellcolor{lightgray}	\underline{0.7698}&\cellcolor{lightgray}0.7244&\cellcolor{lightgray}\textbf{0.7278}\\
&$\rho \uparrow$&\textbf{0.6471}&\underline{0.6528}&\underline{0.6698}&\textbf{0.6929}&\underline{0.5152}&0.6073&\underline{0.5282}&\underline{0.5270}&\underline{0.5812}&0.6301&\textbf{0.5990}&\textbf{0.6100}\\
\hline
\end{tabular}
\caption{Comparison of IQA metrics on handling three challenges. I: Human-object interactions. II: Fantastical ads. III-(P)\&III-(N): Ads designed to evoke positive and negative emotions, respectively. The best two results are highlighted in bold and underlined, respectively.}
\label{tab:challenge1}
\end{table*}
\begin{table*}[t]
\centering
\small
\setlength{\tabcolsep}{1.3mm}
\begin{tabular}{c|c|cccc|cccc|cccc}
\hline
\multirow{2}{*}{Model}                                                     & \multirow{2}{*}{Criteria} & \multicolumn{4}{c|}{Information Clarity} &\multicolumn{4}{c|}{Emotional Interaction}&\multicolumn{4}{c}{Overall}\\
\cline{3-14}
   &                       &I&II&III-(P)&III-(N)&I& II&III-(P)&III-(N)&I&II&III-(P)&III-(N)\\
   \hline
\multirow{2}{*}{LLaVA-v1.5-13B} &\cellcolor{lightgray}$\alpha \uparrow$&\cellcolor{lightgray}0.4864&\cellcolor{lightgray}0.5058&\cellcolor{lightgray}0.4907&\cellcolor{lightgray}0.4733&\cellcolor{lightgray}0.4925&\cellcolor{lightgray}0.5375&\cellcolor{lightgray}0.4944&\cellcolor{lightgray}0.5008&\cellcolor{lightgray}0.4895	&\cellcolor{lightgray}0.5217&\cellcolor{lightgray}0.4926&\cellcolor{lightgray}0.4871 \\
&$\kappa \uparrow$&0.0518&0.0192&0.0343&0.0425&0.4847&0.2558&0.3673&0.4825&0.2683&0.1375&0.2008&0.2625\\
\hline
\multirow{2}{*}{BakLLaVA} &\cellcolor{lightgray}$\alpha \uparrow$&\cellcolor{lightgray}0.5259&\cellcolor{lightgray}0.4933&\cellcolor{lightgray}0.5330&\cellcolor{lightgray}0.4876&\cellcolor{lightgray}0.5111&\cellcolor{lightgray}0.4800&\cellcolor{lightgray}0.5092&\cellcolor{lightgray}0.4876&\cellcolor{lightgray}0.5185&\cellcolor{lightgray}0.4867	&\cellcolor{lightgray}0.5211&\cellcolor{lightgray}0.4876 \\
&$\kappa \uparrow$&0.1704&0.1467&0.1847&0.1570&0.2370&0.1867&0.1741&0.2479&0.2037&0.1667&0.1794&0.2025\\
\hline
\multirow{2}{*}{mPLUG-Owl2} &\cellcolor{lightgray}$\alpha \uparrow$&\cellcolor{lightgray}0.4839&\cellcolor{lightgray}0.4798&\cellcolor{lightgray}0.4857&\cellcolor{lightgray}0.4826&\cellcolor{lightgray}0.4827&\cellcolor{lightgray}0.4673&\cellcolor{lightgray}0.4895&\cellcolor{lightgray}0.4817&\cellcolor{lightgray}0.4833	&\cellcolor{lightgray}0.4736&\cellcolor{lightgray}0.4876&\cellcolor{lightgray}0.4822 \\
&$\kappa \uparrow$&\underline{0.4814}&\underline{0.4952}&\underline{0.4818}&\underline{0.4858}&\underline{0.4919}&\underline{0.4654}&\underline{0.4718}&\underline{0.4865}&\underline{0.4867}&\underline{0.4803}&\underline{0.4768}&\underline{0.4862}\\
\hline
\multirow{2}{*}{InternLM-XC.2-vl} &\cellcolor{lightgray}$\alpha \uparrow$&\cellcolor{lightgray}0.3295&\cellcolor{lightgray}0.3538&\cellcolor{lightgray}0.3127&\cellcolor{lightgray}0.3294&\cellcolor{lightgray}0.4584&\cellcolor{lightgray}0.4500&\cellcolor{lightgray}0.4506&\cellcolor{lightgray}0.4669&\cellcolor{lightgray}0.3940	&\cellcolor{lightgray}0.4019&\cellcolor{lightgray}0.3817&\cellcolor{lightgray}0.3982 \\
&$\kappa \uparrow$&0.2109&0.2654&0.1933&0.2071&0.3058&0.2519&0.2689&0.3038&0.2584&0.2587&0.2311&0.2555\\
\hline
\multirow{2}{*}{IDEFICS-Instruct} &\cellcolor{lightgray}$\alpha \uparrow$&\cellcolor{lightgray}\underline{0.5556}&\cellcolor{lightgray}\underline{0.6133}&\cellcolor{lightgray}\underline{0.5435}&\cellcolor{lightgray}\underline{0.6281}&\cellcolor{lightgray}\underline{0.6815}&\cellcolor{lightgray}\underline{0.6267}&\cellcolor{lightgray}\underline{0.5673}&\cellcolor{lightgray}\underline{0.6364}&\cellcolor{lightgray}\underline{0.6186}	&\cellcolor{lightgray}\underline{0.6200}&\cellcolor{lightgray}\underline{0.5554}&\cellcolor{lightgray}\underline{0.6323} \\
&$\kappa \uparrow$&0.1556&0.2400&0.2005&0.2314&0.3481&0.3333&0.2559&0.5537&0.2519&0.2867&0.2282&0.3926\\
\hline
\multirow{2}{*}{Qwen-VL-Chat} &\cellcolor{lightgray}$\alpha \uparrow$&\cellcolor{lightgray}0.5481&\cellcolor{lightgray}0.5600&\cellcolor{lightgray}0.4987&\cellcolor{lightgray}0.5372&\cellcolor{lightgray}0.3481&\cellcolor{lightgray}0.3200&\cellcolor{lightgray}0.2612&\cellcolor{lightgray}0.2893&\cellcolor{lightgray}0.4481	&\cellcolor{lightgray}0.4400&\cellcolor{lightgray}0.3800&\cellcolor{lightgray}0.4133\\
&$\kappa \uparrow$&0.0006&0.0133&0.0026&0.0083&0.0148&0.0400&0.0158&0.0000&0.0077&0.0267&0.0092&0.0042\\
\hline
\multirow{2}{*}{GPT-4o} &\cellcolor{lightgray}$\alpha \uparrow$&\cellcolor{lightgray}\textbf{0.8296}&\cellcolor{lightgray}\textbf{0.8120}&\cellcolor{lightgray}\textbf{0.7929}&\cellcolor{lightgray}\textbf{0.7944}&\cellcolor{lightgray}\textbf{0.6963}&\cellcolor{lightgray}\textbf{0.7529}&\cellcolor{lightgray}\textbf{0.7207}&\cellcolor{lightgray}\textbf{0.7299}&\cellcolor{lightgray}\textbf{0.7630}&\cellcolor{lightgray}\textbf{0.7825}	&\cellcolor{lightgray}\textbf{0.7568}&\cellcolor{lightgray}\textbf{0.7622} \\
&$\kappa \uparrow$&\textbf{0.8963}&\textbf{0.8947}&\textbf{0.8760}&\textbf{0.7934}&\textbf{0.8889}&\textbf{0.8289}&\textbf{0.8364}&\textbf{0.8760}&\textbf{0.8926}&\textbf{0.8618}&\textbf{0.8562}&\textbf{0.8347}\\
\hline
\end{tabular}
\caption{Comparison of LMMs on handling three challenges. I: Human-object interactions. II: Fantastical ads. III-(P)\&III-(N): Ads designed to evoke positive and negative emotions, respectively. The best two results are highlighted in bold and underlined, respectively.}
\label{tab:challenge2}
\end{table*}
\begin{table}[t]
\centering
\small
\setlength{\tabcolsep}{1.6mm}
\begin{tabular}{c|c|ccc}
\hline
Model    & Dimension& \textit{Comp.}$\uparrow$         & \textit{Prec.}$\uparrow$            & \textit{Rele.}$\uparrow$                            \\
\hline
\multirow{3}{*}{
LLaVA-v1.5-13B}    &     IC                   & 0.8395  & 0.9106& 1.5973 \\
                          & EI                 &   0.6755   &  0.5956  &1.6001\\
                          & \cellcolor{lightgray}Overall                               & \cellcolor{lightgray}0.7575  & \cellcolor{lightgray}0.7531  & \cellcolor{lightgray}1.5987\\
\hline 
\multirow{3}{*}{BakLLava}  &     IC                   & \underline{0.8641} &  \underline{1.0998}& \underline{1.7592}\\
                          & EI                 & 0.6032  & 0.5588 &  \underline{1.6721}\\
                          & \cellcolor{lightgray}Overall    & \cellcolor{lightgray}0.7336  & \cellcolor{lightgray}0.8293 & \cellcolor{lightgray}\underline{1.7157} \\
\hline
\multirow{3}{*}{mPLUG-Owl2}  &     IC                   & 0.7825 & 0.8748 &  1.6835\\
                          & EI                 &\underline{0.7332} &  0.6242  & 1.5910\\
                          & \cellcolor{lightgray}Overall                               &  \cellcolor{lightgray}\underline{0.7579}    &  \cellcolor{lightgray}0.7495& \cellcolor{lightgray}1.6372\\
\hline
\multirow{3}{*}{InternLM-XC.2-vl}  &     IC                   &   0.5903 &0.9094 & 1.5443\\
                          & EI                 &  0.6682 &  \underline{0.7883}  &  1.4601\\
                          & \cellcolor{lightgray}Overall                               & \cellcolor{lightgray}0.6293   &\cellcolor{lightgray}\underline{0.8489} & \cellcolor{lightgray}1.5022\\                          
\hline
\multirow{3}{*}{IDEFICS-Instruct}  &     IC                   &  0.4893  & 0.6082 &   1.4311 \\
                          & EI                 &0.2342 & 0.3513 & 1.1482\\
                          & \cellcolor{lightgray}Overall                               & \cellcolor{lightgray}0.3618 & \cellcolor{lightgray}0.4797 & \cellcolor{lightgray}1.2896   \\
\hline
\multirow{3}{*}{Qwen-VL-Chat}  &     IC                   &  0.7502  &0.7605  &  1.4595 \\
                          & EI                 & 0.5474&   0.4127 &   1.3243\\
                          & \cellcolor{lightgray}Overall                               &  \cellcolor{lightgray}0.6488  & \cellcolor{lightgray}0.5866 &  \cellcolor{lightgray}1.3919  \\       
\hline
\multirow{3}{*}{GPT-4o}  &     IC                   &  \textbf{1.3042}  &  \textbf{1.3974} & \textbf{1.9294}\\
                          & EI                 &  \textbf{1.3504 } &   \textbf{1.6191} & \textbf{1.8981}\\
                          & \cellcolor{lightgray}Overall                               & \cellcolor{lightgray}\textbf{1.3273}   & \cellcolor{lightgray}\textbf{1.5083}    & \cellcolor{lightgray}\textbf{1.9138}  \\      
\hline     
\end{tabular}
\caption{Comparisons of LMMs in \textbf{preference interpretation}. IC: Information clarity. EI: Emotional interaction. The best two results are highlighted in bold and underlined, respectively.}
\label{tab:interpretation}
\end{table}
\begin{table}[t]
\centering
\small
\setlength{\tabcolsep}{1.6mm}
\begin{tabular}{c|c|ccc}
\hline
Model    & Dimension & \textit{Comp.}$\uparrow$         & \textit{Prec.}$\uparrow$            & \textit{Rele.}$\uparrow$     \\
\hline
\multirow{3}{*}{
LLaVA-v1.5-13B}    &     IC                   &0.4006  & 0.2746  &1.7500\\
                          & EI                 &   0.4393&\underline{1.3258}  & 1.6401\\
                          & \cellcolor{lightgray}Overall                               &\cellcolor{lightgray}0.4200    & \cellcolor{lightgray}0.8002   & \cellcolor{lightgray}\underline{1.6951} \\
\hline 
\multirow{3}{*}{BakLLava}  &     IC                   &  0.2848 & 0.4835& 1.6561 \\
                          & EI                 &  \underline{0.4698} & 0.8766 &  \underline{1.6859}  \\
                          & \cellcolor{lightgray}Overall                               &\cellcolor{lightgray}0.3773   & \cellcolor{lightgray}0.6800    & \cellcolor{lightgray}1.6710  \\
\hline
\multirow{3}{*}{mPLUG-Owl2}  &     IC                   & \underline{0.4835}   & 0.3273& \underline{1.7742}\\
                          & EI                 &0.3683  &   0.7191  &1.5609  \\
                          & \cellcolor{lightgray}Overall                               &   \cellcolor{lightgray}\underline{0.4259}   &   \cellcolor{lightgray}0.5232 &\cellcolor{lightgray}1.6676 \\
\hline
\multirow{3}{*}{InternLM-XC.2-vl}  &     IC                   & 0.4524 &  \underline{1.2841} & 1.5978\\
                          & EI                 &  0.3902  &  1.2741&  1.4795 \\
                          & \cellcolor{lightgray}Overall                               &   \cellcolor{lightgray}0.4213 & \cellcolor{lightgray}\underline{1.2791}  & \cellcolor{lightgray}1.5386 \\                          
\hline
\multirow{3}{*}{IDEFICS-Instruct}  &     IC                   &0.4097&  0.4154  & 1.6245\\
                          & EI                 &  0.1685& 0.3305&    1.5738   \\
                          & \cellcolor{lightgray}Overall                               &   \cellcolor{lightgray}0.2891 & \cellcolor{lightgray}0.3729  &   \cellcolor{lightgray}1.5992\\
\hline
\multirow{3}{*}{Qwen-VL-Chat}  &     IC                   & 0.4350  & 1.0615 &  1.6565   \\
                          & EI                 &   0.3935  & 1.1408    &1.6739 \\
                          & \cellcolor{lightgray}Overall                               &\cellcolor{lightgray}0.4142 & \cellcolor{lightgray}1.1011  &\cellcolor{lightgray}1.6652  \\       
\hline
\multirow{3}{*}{GPT-4o}  &     IC                   & \textbf{1.0220}  & \textbf{1.6790}  &  \textbf{1.8232} \\
                          & EI                 &\textbf{0.7910} & \textbf{1.5262} &\textbf{1.8277} \\
                          & \cellcolor{lightgray}Overall                               &\cellcolor{lightgray}\textbf{0.9065} &  \cellcolor{lightgray}\textbf{1.6026} & \cellcolor{lightgray}\textbf{1.8254}  \\      
\hline    
\end{tabular}
\caption{Comparisons of LMMs in \textbf{preference reasoning}. IC: Information clarity. EI: Emotional interaction. The best two results are highlighted in bold and underlined, respectively.}
\label{tab:reason}
\end{table}
\subsubsection{Criteria} We exploit various evaluation criteria to quantify the capabilities of the competing models in terms of preference prediction, interpretation and reasoning. 
Regarding preference prediction, we use three criteria: 1) Correlation ($\rho$): the linear correlation between the ground-truth and predicted preference probabilities; 2) Accuracy ($\alpha$): the ratio of image pairs correctly predicted by the model; 3) Consistency ($\kappa$): the criteria is designed for LMMs, which measures whether the predictions from LMMs are robust to the presentation order of two images. More specifically, Given an image pair $(x,y)$ and its reference information $z$ (text or emotion category). $f$ is the model to be tested, where $f((x,y),z)=1$ if $x$ is preferred over $y$ given $z$, and $f((x,y),z)=0$ otherwise. The accuracy, consistency, and correlation of the model can be computed as follows,
\begin{equation}
    \rho= \text{PLCC}(\mathcal{P}_{(\mathcal{X},\mathcal{Y})|\mathcal{Z}},\hat{\mathcal{P}}_{(\mathcal{X},\mathcal{Y})|\mathcal{Z}}),   
\end{equation}
\begin{equation}
    \kappa= \frac{1}{\mathcal{|D|}} \sum_{((x,y),z)\in \mathcal{D}}\mathbb{I}\left [ f((x,y),z)+f((y,x),z)=1 \right ]  , 
\end{equation}
\begin{equation}
    \alpha= \frac{1}{\mathcal{|D|}} \sum_{((x,y),z)\in \mathcal{D}}\mathbb{I}\left [ f((x,y),z)=\mathbb{I}\left[p_{(x,y)|z}>0.5\right] \right ]   ,
\end{equation}
where $|\mathcal{D}|$ and $\mathbb{I}$ are the total number of pairs and the indicator function, respectively. $p_{(x,y)|z}$ denotes the ground-truth preference probability that $x$ is preferred over $y$ given reference information $z$. $\mathcal{P}_{(\mathcal{X},\mathcal{Y})|\mathcal{Z}}$ and $\hat{\mathcal{P}}_{(\mathcal{X},\mathcal{Y})|\mathcal{Z}}$ represent the ground-truth and the predicted preference probabilities of all pairs in the whole dataset. PLCC is the Pearson linear correlation coefficient measure. It is worth noting that all predicted scores by the model are fitted before computing the preference probabilities. The higher values of $\alpha$, $\rho$, and $\kappa$ signify a better performance of the model.

Regarding preference interpretation and reasoning, we employ the GPT-assisted evaluation method to evaluate LMM responses against the golden descriptions.
Following the suggestions in~\cite{wu2024qbench}, we employ three evaluation criteria: (1) Completeness (\textit{Comp.}): Encouraging LLM outputs that closely align with the golden description; (2) Preciseness (\textit{Prec.}): Penalizing outputs that include information conflicting with the golden description; (3) Relevance (\textit{Rele.}): Ensuring a higher proportion of LLM outputs pertain to information involving the crucial factors of a specific evaluation dimension. 

\subsection{Performance on Preference Prediction}
We input image pairs and their corresponding reference information into the models (except for HyperIQA, as it only supports image inputs) to evaluate performance regarding information clarity and emotional interaction.
For information clarity evaluation, the reference information is the text; for emotion interaction evaluation, the reference information is the emotion category. 
The results are shown in Tables~\ref{tab:preferencepred1}\&\ref{tab:preferencepred2}, where $D_{all}$ and $D_{sub}$ represent all pairs with the full range of preference probabilities from 0 to 1 and a subset of image pairs where humans show strong preferences, respectively. We can see that 1) in terms of prediction accuracy on information clarity and emotional interaction dimensions, GPT-4o significantly outperforms all other competing models, particularly surpassing other LMMs; 2) the prediction accuracy of CLIP-based metrics designed for AIGIs is higher than that of LMMs (excluding GPT-4o) and HyperIQA designed for natural images, indicating that AIGIs present unique characteristics and challenges; 3) all open-source LMMs perform poorly in prediction consistency, suggesting that they tend to provide biased responses regardless of AIGI contents; 4) there is a notable discrepancy in the performance of most models between the information clarity and emotional interaction dimensions, indicating a potential weakness in their ability to assess multiple aspects.

We also design three challenges to compare the performance of the models in handling different contexts. The first challenge focuses on ads with human-object interactions~\cite{jianglin2024record}, such as ``A baby reaching for hanging toys'' and ``A young boy carrying heavy bricks.'' The second challenge centers on fantastical ads, which involve imaginative and visually complex content often featuring surreal or exaggerated elements, such as ``Popcorn kernels forming the shape of a majestic mountain in a natural landscape'' and ``A surreal image of a giant lemon squeezing itself into a tiny bottle.'' The third challenge evaluates the effectiveness of the models on ads designed to evoke positive and negative emotions. The results are shown in Tables~\ref{tab:challenge1}\&\ref{tab:challenge2}, from which one can observe 1) compared to other IQA algorithms, ImageReward excels in information clarity dimension of challenge II, while HPSv2 achieves higher $\lambda$ and $\rho$ values in emotional interaction dimension in challenges I and III; 2) compared to other LMMs, GPT-4o achieves the best performance across these three challenges.

\subsection{Performance on Interpretation and Reasoning}
We evaluate the interpretation and reasoning abilities of the LMMs using golden descriptions. During the interpretation process, the LMMs analyze human choices and infer the reasons behind these preferences. During the reasoning process, we provide two images and require the LMMs to conduct a detailed comparison, ultimately making a preference decision based on the comparative analysis. The results are shown in Tables~\ref{tab:interpretation}\&\ref{tab:reason}. We can draw the following findings: 1) GPT-4o achieves the best performance in preference interpretation and reasoning across all criteria; 2) for both preference interpretation and reasoning, most LMMs exhibit high relevance values but lower completeness and precision values. The results suggest that while LMMs responses effectively address visual cues within each evaluation dimension, they often lack comprehensive coverage and include conflicting information, leading to less accurate and less complete responses.
\section{Conclusion}
In this work, we introduce AIGI-VC, a quality assessment dataset containing 2,500 AIGIs across 14 ad topics and 8 emotion types. AIGI-VC facilitates the quality assessment of AIGIs in terms of information clarity and emotional interaction, providing coarse-grained and fine-grained human preference annotations. Our experimental results highlight the need for an IQA metric to effectively handle the unique characteristics of AIGIs in visual communication. We hope that our dataset and analysis will shed light on the development of more robust and accurate IQA metrics, enhancing the effectiveness of AIGIs in practical applications.

\section{Acknowledgments}
This work is partially supported by the Science and Technology Innovation
2030 Key Project (Grant No. 2018AAA0101301), the Hong Kong Innovation
and Technology Commission (InnoHK Project CIMDA), and the
Hong Kong General Research Fund under Grant 11209819, 11203820, 11200323 and 11203220.

\bibliography{aaai25}
\end{document}